# REAL TIME DEEP LEARNING WEAPON DETECTION TECHNIQUES FOR MITIGATING LONE WOLF ATTACKS

Akhila Kambhatla[1] and Ahmed R Khaled[2]

[1]School of Computing, Southern Illinois University, Carbondale, Illinois, USA
akhila.kambhatla@siu.edu
[2]School of Computing, Southern Illinois University, Carbondale, Illinois, USA
kahmed@cs.siu.edu

## ABSTRACT

Firearm Shootings and stabbings attacks are intense and result in severe trauma and threat to public safety. Technology is needed to prevent lone-wolf attacks without human supervision. Hence designing an automatic weapon detection using deep learning, is an optimized solution to localize and detect the presence of weapon objects using Neural Networks. This research focuses on both unified and II-stage object detectors whose resultant model not only detects the presence of weapons but also classifies with respective to its weapon classes, including handgun, knife, revolver, and rifle, along with person detection. This research focuses on $YOLOv_5$ (You Look Only Once) family and Faster RCNN family for model validation and training. Pruning and Ensembling techniques were applied to *YOLOv5* to enhance their speed and performance. $YOLOv_5$ models achieve the highest $F_1$ score of 78% with an inference speed of 8.1ms. However, Faster R-CNN models achieve the highest *AP* 89%.

## 1. INTRODUCTION

Most deaths globally involve weapons which have a traumatic impact on health and psychological and economic costs. According to Gun Violence Archive, 44266-gun violence deaths are recorded [1] in the United States, which would cost around $ 557 billion as an economic consequence [2]. To achieve peace and enhance safety, it is highly required to reduce gun violence globally. Governments and private security units have been expanding the use of surveillance cameras in public places like airports, banks, and public gathering events which supports detecting intruders, crime investigation, and proximity alarms and even serve as evidence for judiciary courts. The two limitations of surveillance cameras are: a) they cannot detect the attack in advance b) they need continuous human monitoring, which oscillates concentration at the rate of 83%, 84%, and 64% after one hour of continuous monitoring 4,9,16 screens respectively [3].

Weapon detection is mainly categorized into traditional & [4, 5, 6, 7, 8], image processing techniques [9, 10, 11, 12, 13], Machine Learning (ML) algorithms [14, 15, 16] and deep learning algorithms [17, 18, 19, 20, 21, 22, 23]. Traditional weapon detection techniques mainly focus on thermal/infrared and X-Ray techniques for detection are expensive and excess usage leads to radiation which cause deadly diseases like cancer and malignant tumours [24]. Several techniques have been employed to detect concealed weapons such as pattern matching, density descriptors, image segmentation, and cascade classifiers, which produce false alarms [13]. Research efforts used RGB surveillance images [25] and applied image processing techniques to detect weapons in the shooting scenes such as Image Fusion [9], colour segmentation [11], Harris interest points [14], and Speeded-Up Robust Features.

Weapon detection using ML techniques involves three phases which are explained in detail in the next section [II]. Some of the pragmatic approaches in ML are the Haar-like features [26], Histogram of Gradients [14], Bag of words [15, 16] and Scale Invariant Feature Transform [26] were used to identify feature vectors from an input image which may have reduced human contribution to a certain extent. Some of the stated image processing techniques combined with machine learning algorithms such as support vector machine (SVM) [27] to classify the detected weapons. The use of ML techniques improves classification accuracy.

However, the limitations include much computational time and sophisticated engineering design in feature extraction. However, Deep Learning (DL) techniques employ neural networks to extract features with multiple levels of abstraction for overcoming the limitations of ML algorithms for object detection [28]. Neural networks are mainly divided into two groups: (1) One-Stage object detectors or unified networks as You Look Once Only (YOLO) [29, 30, 31, 32, 33] (2) Two-Stage Object Detectors as Region-based Convolutional Neural Network (RCNN) [34, 35, 36]. Unified networks are known for their speed and no need for cascading area classification. Two-Stage detectors initially produce region proposals and extract features from each proposal region showed promising results as shown in [17, 18, 19, 20, 21, 23, 22]. However, researchers are still eager to generate a DL model with adequate accuracy and fast enough to prevent mass shootings and stabbings attacks in real-time by automating the detection of several weapons classes in the market. The model evaluation and performance might be impacted due to having some inactive neurons that have less usage in the training model.

The main objective of this research study is to identify research gaps in the weapon detection field and overcome its limitations. The main aim is to examine the strength and shortcomings of the unified and two-stage object models with the unique weapon dataset that classifies five classes and offers a detailed assessment of pragmatic results in terms of accuracy, precision, and recall with open issues with forecasting of future aspects. The main contributions of this paper are summarized as follows:

- We explored different object detection models, among unified and two-stage object detectors to determine the better generic weapon design model to detect and classify small and faraway weapon objects.

- The ImageNet [37] dataset could be used to classify and detect only two types of weapons: pistol and revolver. In this paper, we collected and labelled several images, including different types of weapons to be classified such as handgun, knife, revolver, and rifle, along with human detection.

- The original model was developed, validated, pruned, and combined with different models from the same family for detecting weapons.

In summary, from *YOLOv5* family, $Y_s, Y_m, Y_l$ are used with $C_3 - SPP$ backbone. For feature extraction using two-stage weapon detectors *Resnet 50, Resnet 101,* models are used with backbones $FPN, C_4, DC_5$ respectively. *Retinanet 50 and Retinanet 101* with FPN backbone network are used to detect weapons in this research.

The paper is organized as follows: the related work is briefly introduced in section [II]. Section [III] comprises the architectural view of the YOLOv5 family and Faster R-CNN, along with dataset collection and metric evaluation. Environmental setup, experiments, and results are presented in section [IV]. Conclusions and further enhancements are in mentioned in section[V].

## 2. RELATED WORK

All The detection and classification of weapons approaches have been divided into Image Processing techniques, Machine Learning Algorithms, and model-based approaches by Deep Learning.

### 2.1. Image Processing & Traditional Weapon Detection Techniques

Transportation Security Administration (TSA) recorded 6600 weapons discovered in carry-on bags at checkpoints and the secure area of airports across the USA in 2022, which is a 10% increase from 2021 [38]. The Microwave swept frequency radar [4] technique detects metallic objects by reflection technique. Metallic objects are distinguished in terms of swept frequency response, including the object's shape, position, and frequency behaviour. Moreover, CWD techniques use different combinations to achieve higher accuracy by Pattern Matching [7], Density Descriptors [8], Image Segmentation [6], and Cascade Classifiers with Boosting [5]. CWD approaches are limited in detecting metallic objects. Non-metallic weapons cannot be detected. Deploying X-ray scanners and Conveyor belts on a large scale is expensive. Darker proposed a complex automated real-time analysis software application [10] called Multi Environment Deployable Universal Software Application (MEDUSA) that recognizes gun crimes from Closed-circuit television (CCTV) footage which need human monitoring and expensive in maintenance. Image fusion is an image processing approach for detecting weapons or objects hidden beneath a person's clothing [11] by IR (Infrared) algorithmic approach. A hybrid weapon detection using fuzzy logic was developed to reduce false alarm rates using a combination of metallic test and image test arms [13]. To detect weapons in RGB images, Tiwari and Verma [25] used colour segmentation, SIFT [39] and k-means algorithm [40] to remove unrelated objects. Then, they applied the Harris interest point detector and Fast Retina Key point (FREAK) [41] to accurately locate the handgun in the segmented images 84.26%. Speeded-Up Robust Features (SURF) method also used the K-Means clustering algorithm to detect pistols from segmented images and achieved accuracy 88.67% [42]. The limitation of this approach is a spatial distortion where the visual image may be distorted, producing false negative cases while detecting weapons.

### 2.2. Machine Learning Methods

Object Detection is one of the famous and challenging techniques to detect objects such as weapons and aims to locate and classify existing real-world objects in any image and label them in rectangular bounding boxes with respective class labels along with confidence scores [43]. These detection techniques also handle occlusions, variations, transformations, and degradation without losing the accuracy of the classification. The main aim is to find the object zone through multi-scale windows. The second stage is to extract a feature vector of fixed length and secure the specific information of the area enclosed. Finally, area classifiers were trained to identify classes. Applying ML algorithms to low-level visual descriptors such as Histogram of Oriented Gradients (HOG) [14], bag-of-words [15, 16] and Haar-like features [26] improve object detection model as a classification problem with significant improvements such as feature extractions. Additional methods for classification approaches, such as bagging [44], cascade learning [45], and AdaBoost [46], were better in resulting in detection accuracy. However, the machine learning algorithms require high computational time, cannot scale on extensive data, and have high chances of model overfitting. If ML algorithms focus on automatic feature extractions through neural networks, then the area is termed as deep learning.

### 2.3. Evolution of Deep Learning Methods

Weapon detection using CNN uses two approaches: sliding windows and region proposal [47]. The sliding windows approach uses a region with multiple scales to scan the input image at all locations and then apply the classifier at each one. However, the region proposal approach considers all candidate regions instead of all windows in the sliding window approach. In [19] used Faster R-CNN to detect guns using Google Net and knives using Squeeze Net backbone networks and achieved mean Average Precision (mAP@0.50) 85.4% and 46.68% respectively. Alaqil et al. [20] used Faster R-CNN to detect guns using Inception-ResNetV2 and VGG16 and achieved mAP@0.50 81%, and 72% respectively. The authors in [47] trained Faster R-CNN using VGG16 classifier on ImageNet [48] and then fined tuned on their dataset of high-quality gun images. To achieve real-time recently, the authors in [22] proposed a weapon detection system using YOLO trained on images obtained from Internet Movie weapon Database (IMFDB) [49]. Their approach achieved 96.26% accuracy and 70% of mAP in detecting weapons. In our previous research in the Big Data Software Engineer (BASE) research lab at Southern Illinois University [17], we implemented the sliding window approach and region proposal approach by two deep neural models SSD [50]and Faster R-CNN [51].

We concluded that SSD fails to detect smaller weapons on the scene and is known for its speed and fast performance. In contrast, the Faster R-CNN model took a long computational time to detect weapons with higher accuracy. In conclusion, research on weapon detection in images or videos is relatively sparse, and currently, there is no dedicated weapon detector or weapon benchmark dataset for performance evaluation and comparison. This research aims to develop and generate a training model to detect and classify all the weapon categories accurately and in real time. All the previously mentioned models failed to detect small and faraway weapons objects. Therefore, we developed the YOLOv5 and Faster R-CNN, a real-time object detector whose network architecture differs with each backbone. YOLOv5 is known for its inference speed and accuracy for its average computation time. On the other hand, Faster R-CNN is known for its accuracy and generates more true positives.

## 3. METHODOLOGY

The methodological approach is to determine the better object detection model for weapon detection using Supervised Learning. Object detection mainly focuses on bounding box detection, where annotations are annotated manually by marking the axis aligned in rectangular boxes.

### 3.1. Dataset Collection & Data Augmentation

Many famous datasets like PascalVOC, ImageNet, and Open Image Datasets categorize weapons as a single class rather than classifying them into different categories. In many cases, weapons are classified as either pistols or guns but not classified as it is typed. The dataset images are collected from Google and a collection of frames from live videos from YouTube. As precisely, the focus of the dataset is on the type of gun, thereby classifying the gun as a rifle, handgun, revolver, and knife, along with human detection. Data annotation and Labellmg are done manually and in three steps. Initially determining the classes for Labellmg, Collecting the images, and dividing them for training, testing and validation. Labellmg is a graphical image annotation tool used in the research whose interface is python, in which annotations are saved in YOLO format, which is .txt files. The homogeneous dataset consists of 2500 samples comprising 500 @ of each class.

Data Augmentation is one of the solutions for low-resolution images where the image is subjected to various transformations such as cropping, flipping, rotating, blurring, scaling, translating, colour perturbations, and adding noise. In addition, it increases the amount of training and validation datasets to improve the ability of generalization in deep-learning models. Figure [1] depicts the original image 1(b) and the image after augmentation 1(a). Image augmentation algorithms include geometric transformations, colour space transformations, kernel filters, feature space augmentation techniques, neutral style transfer, and meta-learning [52]. In this paper, we used the "Imguag" [53] python library to expand an existing image dataset by artificially creating variations in existing images. The entire dataset includes images of 640 × 640 pixels. Mosaic augmentation was applied in Yolov5 [54] where four or more random images are combined into four tiles with a random ratio with the relative scale of the image and trained through a data loader.

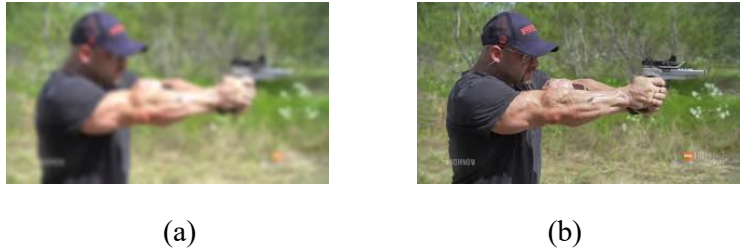

(a)                (b)

Figure 1. Data Augmentation Sample after blurring (a) and original Image (b)

## 3.2. Model Analysis & Network Architecture:

Generalization explains the difference in model performance on training and testing data evaluation. The model evaluation is successful if validation error is reduced to decrease with training error. Generic Object detectors were classified into two categories: One-stage and Two-stage detectors, where a model is trained and validated for the detection rate of weapons.

### 3.2.1. One-Stage Weapon Detector

YOLOv5 is a unified object detector known for its speed and performance. YOLOv5 combines initial YOLO to YOLOv4, where continuous performance and structure improvements are made. The original YOLO is a regression problem where the input image is divided into $S \times S$ grids where each grid is responsible for detecting the presence of an object (e.g., weapon). There were β bounding boxes, and confidence scores were predicted for the cells only if there was any weapon presence. Every grid cell supports multiple bounding boxes, and prediction is of 5 attributes $x, y, w, h, c$ where $x, y$ are centre of the bounding box, $w, h$ are width and height of an image and c represents confidence score. Therefore, there will be $S * S * \beta * 5$ outputs for a single input image. Depending upon the confidence score, one can be sure of the presence or absence of a weapon. The major changes of Yolov5 architecture made in this research are:

1. In the backbone, CSPNet [55] is replaced with [56] Neural Network (NN)

2. There is no specific neck in the structure. However, Bottle Neck NN is used within the C3 NN structure.

3. ReLU [57] activation is replaced with SiLU [58] (Sigmoid Linear Activation) within every Convolutional layer.

There are a total of 24 convolutional layers in the YOLOv5 Neural network, starting with the focus layer, backbone (9), and head (14) which is depicted in figure [2] [59]. The model backbone is used to extract essential features from an input image. Initiating the first layer, the focus which takes the input as tensors and expands it into channel space (c-space) and ends with the eighth layer, which is Spatial Pyramid Pooling (SPP) [60] resulting in 1024 dimensions. SPP is prevalent in generating fixed-length windows irrespective of input image size. For better weapon detection, max pooling is done with selected kernel lengths among 5, 9, and 13. The output of the VIII-layer dimension is 1024, where it is pruned to half in C3 NN. The IX-layer acts like a bridge between the backbone and the head. Bottleneck from C3 NN is used to generate feature pyramids which helps the model to generalize on object scaling. It also helps the exact object identification with different scales and sizes. The head model consists of 14 Convolutional Layers, and the final layer is the Detection layer, the Conv2d network layer. In weapon detection, layers chosen for final detection are 17, 20, and 23. The output of these layers consists of different pixel ranges from 256 to 1024 and is sent to detect the object based on the number of classes with the kernel stride. The model head applies anchor boxes on features and generates final output vectors with class probabilities, objectness scores, and bounding boxes.

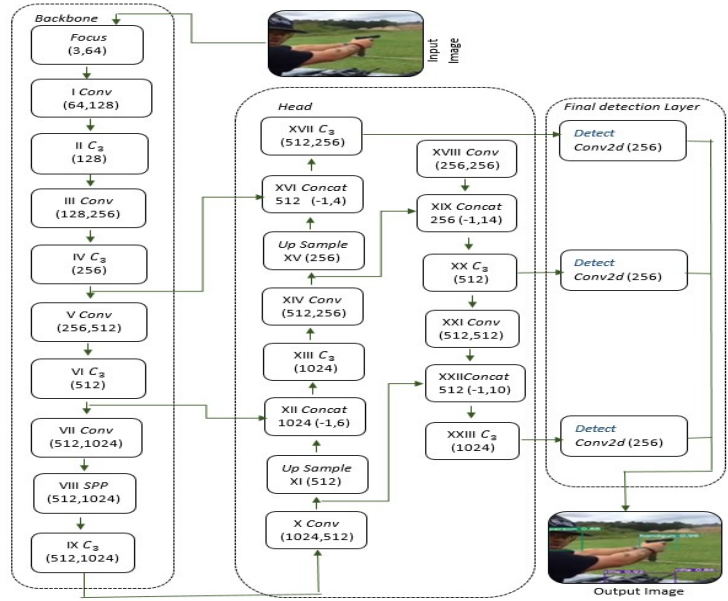

Figure 2: *YOLOv5* Architecture [17]

### 3.2.2. Two-Stage Weapon Detector

Faster R-CNN which is two-stage object detectors consists of two modules a) Region proposal network (RPN), which extracts features and object scores. b) Regions of Interest (RoI) which classifies the above features into object classes with confidence scores. The output from the last conv layer is sent to a small network RPN, which takes the n * n spatial window (n= 3) of the feature map. Later passed through two connected layers: one for regression and the other for classification. Therefore, for the convolutional feature map of size W * H (width * height), there are WHk boxes where k represents the anchor in predicting multiple regions proposals. Translation-Invariant property in anchor explains that any proposal translates an object in an image, and the same function should be able to predict the proposal at either location. The advantage of this is the reduction of model size and dimensionality. There are nine default anchors present on each region proposal. The default anchor sizes are 128,256,512, which can be overridden and observed in figure [3]. Initially, the binary label is assigned to each anchor to determine the presence of an object. The anchor with the highest IoU (Intersection over Union) overlaps with a ground truth box or the anchor with IoU greater than 0.7 is termed a positive label, and the IoU with less than 0.3 is termed a negative label. The overall loss of the RPN is the combination of classification and regression loss.

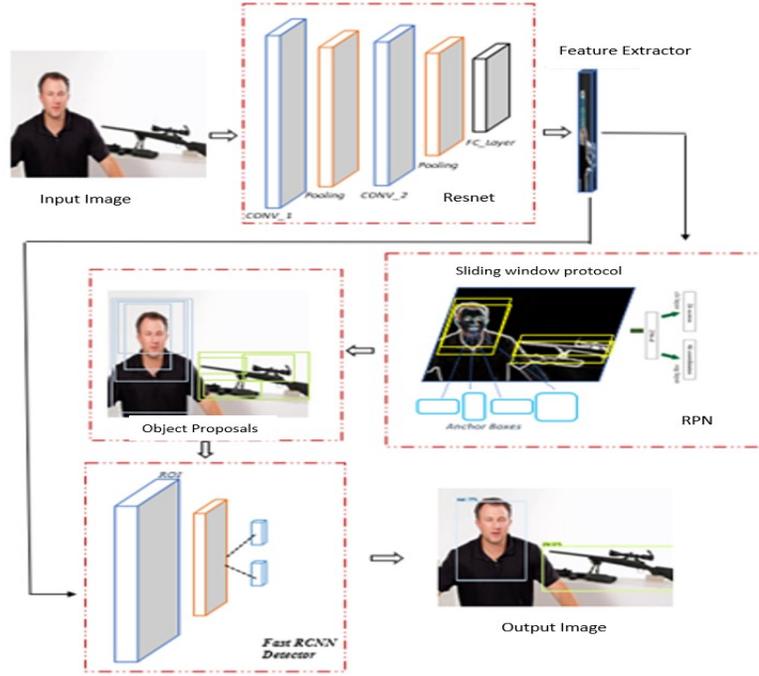

Figure 3: Faster RCNN Architecture [17]

### 3.3. Environment Setup & Evaluation Metrics:

All models are built on NVIDIA GeForce RTX 2060 SUPER, with i5 9^th gen Intel core with 8 GB RAM. Anaconda was used to manage and deploy python 3.8, Pytorch [61], and other essential python libraries. Detectron2 is used to train Faster R-CNN, an open-source object detection platform from Facebook AI. The model can be evaluated with many statistical approaches such as binary cross-Entropy, F1 curve, Precision, Recall, and Accuracy, which are depicted concerning datasets. A good object detection model has both high precision and recall. F1-score is the harmonic mean of precision and recall. The metrics discussed are formulated by equations below Eq (1..6).

$$Recall\ (R) = \frac{TP}{TP+TN} \quad (1)$$

$$Precision\ (P) = \frac{TP}{TP+FP} \quad (2)$$

$$Accuracy = \frac{TP+TN}{TP+TN+FP+FN} \quad (3)$$

$$F_1 Score = \frac{2*P*R}{P+R} \quad (4)$$

$$mAP = \int_0^1 p(r)dr \quad (5)$$

$$IoU = \frac{A \cap B}{A \cup B} \quad (6)$$

a) $TP\ \&\ TN$: True Positives and True negatives are the outcomes where the model correctly predicts the positive and negative weapon class.

b) $FP\ \&\ FN$: False Negatives and False positives are the outcomes where the model incorrectly predicts the positive and negative weapon class.

### 3.4. Model Ensembling & Pruning

This research focuses on averaging multiple models and unstructured pruning over unified networks in improving detection accuracy and to reduce the variance of predictions and reduce generalization error. The reasons for the success of ensemble learning include statistical, computational, and representational learning, along with strong correlation [62]. The technique is used by varying combinations by weighting the predictions of each model over a validation dataset. In this research, trained models $(Y_s, Y_m, Y_l)$ are ensembled together validation where final prediction enhances the model accuracy as displayed in tab [2]. The metric used in this ensemble learning is formulated below Eq (7):

$$Model\ Ensemble = \frac{\sum Model\ Weights}{no:\ of\ models\ ensembled} \quad (7)$$

Pruning is a deep learning technique aims to optimize the model by trimming or eliminating the values of the weight tensors among convolutional layers to save time and resources during the execution of the models. Deep Convolutional neural networks suffer from computational complexity and dimensionality reduction error. Unstructured pruning aims to reduce the number of parameters and operations involved in the computation by removing connections between the neural layers. However, model pruning should be done cautiously as it will permanently damage neural networks. Pruning takes k% sparsity and this research focuses on pruning the YOLOv5 model @30%.

## 4. EXPERIMENTS & PRAGMATIC RESULTS

### 4.1. Dataset & Hyperparameters Tuning

Images are collected, argumentative and scaled at $640 \times 640$ pixels and comprises approximately 2500 images with 500 samples of each category- knife, handgun, revolver, rifle, and person. Noise is added to the dataset as random images, which has the most adverse effect on the accuracy dataset [63]. The homogenous dataset is divided as (75%) and (25%) for training validation respectively. Testing is performed on real-time crime videos. Hyper Parameters plays a crucial role where every model has about at least 20+ hyperparameters for training. Some of them are learning rate, optimization, epochs, and batch size, which should be unique; otherwise, the resulting model consequences could be adverse. The default fitness function is used as a weighted combination of metrics: *mAP@0.5* contributes 10% of the weight, and *mAP@0.5:0.95* contributes the remaining 90%, with Precision P and Recall R absent.

All the unified models are trained up to 1200 epochs. For the YOLOv5 $Y_l$ model, the batch size is reduced from 64 to 40 as the model encountered memory allocation and assertion errors. Faster R-CNN experiments are iterated to 18,000 iterations, and the best results are generated from varying learning rates from 0.00033 - 0.00041, which depends on the neural network model. Resulting models from the *YOLO* family are ensembled to aggregate the prediction and reduce the final prediction's generalization error for unseen data. The stochastic Gradient Descent (SGD) approach is used for optimization.

## 4.2. Pragmatic Analysis

This section presents the performance metrics of YOLOv5, Faster R-CNN models to detect and classify weapons. Unified network models training comprises original training models $Y_s, Y_m, Y_l$, the pruning original models @30% $Y_{sp}, Y_{mp}, Y_{lp}$ and ensemble models $Y_{sm}, Y_{ml}, Y_{ls}$. Eight Faster R-CNN family models are trained and validated to determine the better weapon detection model. Backbones $FPN, C_4, DC_5$ are selected with Resnet and Retinanet with (50, 101) dense-level network layers.

### 4.2.1. One-Stage Weapon Classification Results

The performance of the three original YOLOv5 models $Y_{so}, Y_{mo}, Y_{lo}$ in detecting and classifying weapons along with pruning and ensembling are shown in table [1]. Precision, recall, $mAP@0.5$, $mAP@0.95$ along with the area curves of $F_1$, P, R, PR are the metrics used to analyse the performance of the generated weapons detection models and to determine the best model among them. Figure 10(a) explains mean average precision curves of all YOLOv5 models where ranking is $Y_{so} > Y_{mo} > Y_{lo}$ from highest to least. Figure 10(b) depicts the classification loss of the Yolov5 model ranks $Y_{so} < Y_{mo} < Y_{lo}$ from least to highest. $Y_{so}$ model shows the best precision with 89.4% whereas $F_1$ curve shows no change and has 78% is recorded among $Y_{so}, Y_{sp}, Y_{so.mo}, Y_{lo.so}, Y_{so.mo.lo}$ models. Loss values for the graph $Y_s, Y_m,$ and $Y_l$ models show the same pattern curve as shown in figure 10(b) where $Y_l$ has larger training and validation error of around 0.02 in terms of bounding box, classification and object.

*Table 1: YOLOv5 Metric Comparisons between Original, Prune and Ensemble models.*

| Parameters (%) | Original Model | | | Pruning Model | | | Model Ensemble | | | |
|---|---|---|---|---|---|---|---|---|---|---|
| | $Y_{so}$ | $Y_{mo}$ | $Y_{lo}$ | $Y_{sp}$ | $Y_{mp}$ | $Y_{lp}$ | $Y_{so},Y_m$ | $Y_{mo},Y_{lo}$ | $Y_{lo},Y_{mo}$ | $Y_{so},Y_{mo},Y_{lo}$ |
| Inference Time (s) | .0081 | .019 | .029 | .0029 | .019 | .029 | .0093 | .015 | .016 | .018 |
| Precision | 89.4 | 83.5 | 83.0 | 75.4 | 85 | 78.5 | 85.2 | 82.9 | 85.6 | 80.5 |
| Recall | 70.1 | 68.6 | 67.6 | 70.2 | 62.7 | 64.10 | 72.5 | 69.2 | 72.8 | 76.1 |
| mAP@0.5 | 80.5 | 73.2 | 75.7 | 75.5 | 70.4 | 73.6 | 81.8 | 75.3 | 81.7 | 81.8 |
| mAP@0.95 | 51.3 | 51.2 | 52.3 | 44.0 | 49.4 | 48.9 | 54.4 | 51.8 | 53.9 | 54.7 |
| $F_1$ Score | 78.0 | 74.77 | 74.0 | 78.0 | 74.7 | 74.0 | 78.0 | 75.0 | 78.0 | 78.0 |
| P Curve | 90.5 | 95.8 | 96.0 | 90.0 | 95.8 | 96.0 | 95.0 | 96.4 | 96.4 | 96.0 |
| R Curve | 94.0 | 76.0 | 76.0 | 90.0 | 95.8 | 96.0 | 95.0 | 81.0 | 94.0 | 95.0 |
| PR Curve | 80.7 | 73.2 | 75.0 | 80.0 | 73.2 | 75.0 | 81.8 | 75.3 | 81.7 | 81.8 |

Table [2] shows the performance of classifying individual classes. It shows that $Y_{so}$ model achieved higher precision (94.1%) for classifying Knife and $Y_{lo}$ model achieved precision for classifying rifle class (66.3%). Considering the Recall metric for individual class detections, $Y_{mo}$ has scored more than 84.1% in classifying handgun whereas $Y_{so}$ scored 46.1% in classifying rifle. Pruned YOLOv5 models are depicted in fig [6], have achieved the shortest inference time of about 0.0029 seconds which is amazingly fast in detecting weapons, however, they achieved the least precision in comparison with original YOLOV5 models. Original YOLOV5 models are also ensembled to enhance overall model performance w.r.t weapon detections, metric comparisons, and class detections as shown in figure [5] and tables [1&2]. The ensemble Yolov5 model $Y_{so}, Y_{mo}, Y_{lo}$ has achieved the highest $mAP@0.95$ of 81.8% with the longest inference time of 0.018 seconds.

Table 2: YOLOv5 Class Detections with respective to Precision, Recall and mAP

| Class | Original model | | | Pruning @ 30 % | | | Model Ensemble | | | |
|---|---|---|---|---|---|---|---|---|---|---|
| | $Y_{so}$ | $Y_{mo}$ | $Y_{lo}$ | $Y_{sp}$ | $Y_{mp}$ | $Y_{lp}$ | $Y_{so},Y_{mo}$ | $Y_{mo},Y_{lo}$ | $Y_{lo},Y_{mo}$ | $Y_{so},Y_{mo},Y_{lo}$ |
| | | | | | | Precision | | | | |
| Handgun | 93.5 | 87.9 | 89.9 | 76.1 | 87.7 | 91.5 | 88.5 | 89.7 | 89.9 | 85.7 |
| Knife | 94.1 | 87.9 | 86.0 | 86.8 | 90.4 | 79.9 | 90.7 | 88.9 | 92.6 | 86.5 |
| Person | 85.6 | 81.8 | 80.4 | 68.2 | 78.9 | 76.0 | 82.6 | 78.8 | 82.2 | 78.1 |
| Revolver | 88.1 | 85.2 | 86.2 | 79.7 | 89.9 | 78.5 | 77.6 | 87.6 | 82.3 | 81.2 |
| Rifle | 85.7 | 74.5 | 72.6 | 66.3 | 78.0 | 66.6 | 78.8 | 69.7 | 81.1 | 71.0 |
| | | | | | | Recall | | | | |
| Handgun | 82.9 | 84.1 | 81.7 | 87.8 | 81.7 | 78.4 | 84.1 | 84.9 | 84.1 | 87.5 |
| Knife | 67.6 | 61.5 | 64.8 | 55.6 | 50.7 | 50.3 | 68.7 | 67.6 | 70.4 | 72.5 |
| Person | 80.4 | 78.2 | 71.5 | 84.8 | 65.2 | 67.4 | 82.4 | 72.8 | 82.6 | 85.2 |
| Revolver | 73.5 | 71.4 | 71.4 | 71.4 | 71.4 | 70.4 | 77.6 | 79.5 | 75.5 | 79.5 |
| Rifle | 46.1 | 47.9 | 48.7 | 51.3 | 44.4 | 53.8 | 49.6 | 47.2 | 51.2 | 55.6 |
| | | | | | | mAP @ 0.5 | | | | |
| Handgun | 92.5 | 89.0 | 89.7 | 87.1 | 87.8 | 88.9 | 92.8 | 90.2 | 93.3 | 93.0 |
| Knife | 84.5 | 69.0 | 74.7 | 69.8 | 60.3 | 65.0 | 84.3 | 74.8 | 85.1 | 84.2 |
| Person | 84.1 | 78.3 | 76.7 | 83.4 | 74.6 | 74.2 | 84.7 | 76.6 | 82.2 | 83.6 |
| Revolver | 79.7 | 76.6 | 79.6 | 78.6 | 76.6 | 79.4 | 84.9 | 79.5 | 83.9 | 85.4 |
| Rifle | 61.8 | 53.1 | 57.6 | 58.7 | 52.7 | 60.6 | 62.3 | 55.7 | 64.0 | 62.9 |
| | | | | | | mAP @ 0.95 | | | | |
| Handgun | 71.0 | 72.8 | 73.9 | 62.6 | 71.8 | 67.7 | 73.7 | 73.9 | 72.8 | 74.0 |
| Knife | 55.4 | 43.4 | 45.9 | 39.5 | 39.2 | 40.3 | 53.4 | 46.6 | 52.2 | 51.7 |
| Person | 45.9 | 47.0 | 45.1 | 39.6 | 44.9 | 43.8 | 49.3 | 44.2 | 48.1 | 48.3 |
| Revolver | 51.4 | 58.6 | 59.2 | 49.3 | 58.4 | 55.9 | 59.6 | 58.7 | 61.8 | 61.7 |
| Rifle | 32.7 | 34.1 | 37.3 | 29.2 | 32.7 | 36.4 | 35.1 | 35.4 | 36.7 | 36.5 |

### 4.2.2. Two-Stage Weapon Classification Results

The best Faster RCNN performance results are tabulated in table [3], and the average precision of each weapon class are tabulated in table [4]. Resnet 50 FPN backbone shows the best average precision 89.0% with Intersection over Union $(IoU) = 0.50$ with an inference time of 0.07 seconds. Resnet 101 with $R-10-C_4$ backbone is the slowest model with inference time of 0.26 seconds. Retinanet 101 with R-101-FPN backbone achieves the highest Average precision $(IoU) = 0.50$ in classifying handgun and revolver with 78.03% and 78.02%, respectively.

Table 3: Results of Faster R-CNN Models

| Model | Backbone | LR | AP @ 0.50 | AP 0.5-0.95 | Recall | Inference Time(s) | $F_1$ Score |
|---|---|---|---|---|---|---|---|
| | R-50-FPN | 0.00036 | 89.0 | 56.2 | 63.8 | 0.07 | 74.32 |
| Resnet 50 | R-50-$C_4$ | 0.00041 | 86.97 | 58.7 | 63.1 | 0.26 | 73.13 |
| | R-50-$DC_5$ | 0.00039 | 86.69 | 58.6 | 63.9 | 0.12 | 73.57 |
| | R-101-FPN | 0.00041 | 85.6 | 61.1 | 64.9 | 0.09 | 73.82 |
| Resnet 101 | R-101-$C_4$ | 0.00041 | 87.04 | 59.5 | 63.3 | 0.37 | 73.29 |
| | R-101-$DC_5$ | 0.00033 | 85.36 | 54.3 | 63.4 | 0.16 | 72.25 |
| Retinanet 50 | R-50-FPN | 0.00036 | 86.54 | 56.4 | 64.6 | 0.107 | 73.97 |
| Retinanet 101 | R-50-FPN | 0.00036 | 87.74 | 61.4 | 65.0 | 0.09 | 74.67 |

Considering the size of the generated models, Resnet 50 with backbone R-50-FPN is the lightest model with 283.57Mb while Resnet 101 with backbone R-101-$DC_5$ is the heaviest model with 1438.69 Mb. The limitation of the Faster R-CNN model is that they take much more computational time than unified models, but accurate detection and classifications are reversed. Figure [10c, 10d] depicts the Average precision (AP) and classification loss. Overall, eight models from Faster RCNN have performed similarly with slight variations, which are described in table [3]. In comparing the AP from table [4], The difference between the highest performance model Resnet 50 and least performance model Resnet 101 with same backbone architecture FPN is about 3.4% overlapping difference. In order to avoid such circumstances, a feasible solution is described in the next section.

Table 4: Faster RCNN Average precision IoU@0.50 individual Class Detections Results

| Model | Backbone | LR | AP @ 0.50 | AP @ 0.5-0.95 | Recall | Inference Time(s) | $F_1$ Score |
|---|---|---|---|---|---|---|---|
| Resnet 50 | R-50-FPN | 0.00036 | 89.0 | 56.2 | 63.8 | 0.07 | 74.32 |
| | R-50-$C_4$ | 0.00041 | 86.97 | 58.7 | 63.1 | 0.26 | 73.13 |
| | R-50-$DC_5$ | 0.00039 | 86.69 | 58.6 | 63.9 | 0.12 | 73.57 |
| Resnet 101 | R-101-FPN | 0.00041 | 85.6 | 61.1 | 64.9 | 0.09 | 73.82 |
| | R-101-$C_4$ | 0.00041 | 87.04 | 59.5 | 63.3 | 0.37 | 73.29 |
| | R-101-$DC_5$ | 0.00033 | 85.36 | 54.3 | 63.4 | 0.16 | 72.25 |
| Retinanet 50 | R-50-FPN | 0.00036 | 86.54 | 56.4 | 64.6 | 0.107 | 73.97 |
| Retinanet 101 | R-50-FPN | 0.00036 | 87.74 | 61.4 | 65.0 | 0.09 | 74.67 |

## 4.3. Discussion & Comparison

The results from eighteen models enable us to rank the models based on the inference time and show that $Y_s$ is the fastest model and $Y_l$ is the slowest one in YOLO family as shown in table [1]. The highest $F_1$ score value irrespective of the used model from *YOLOv5* family is 78% followed by $Y_m$ and $Y_s$. However, the $Y_l$ succeeded to detect the smallest number of weapons and ranked as follows $Y_l < Y_m < Y_s$. *Retinanet 101FPN* has achieved the highest recall of $65.0\%$ and $F_1$ score of $74.67\%$. *Retinanet 50 FPN* has above average results as shown in table [4]. In summary, $FPN$ backbones achieved better performance than $DC_5$ and $C_4$. In figure [10], unseen images were tested by the models in which Faster R-RCNN models can able to detect all of the weapon classes [Fig:7(a-c), 8(a-c),9(a,b)]. However, $Y_{ma}$ model cannot be able to detect the knife class fig[6b], while $Y_{sa}$ models detect all classes in fig[6a]. We recommend using *ResNet-50-FPN* model as it achieves the trade-off model and balances the cost in detecting and classifying weapons.

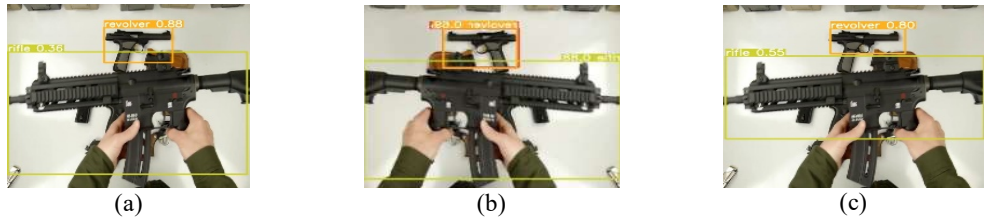

(a)       (b)       (c)

Figure 4: Sample YOLOv5 detections.
(a) $Y_{sa}$ (b) $Y_{ma}$ (c) $Y_{la}$

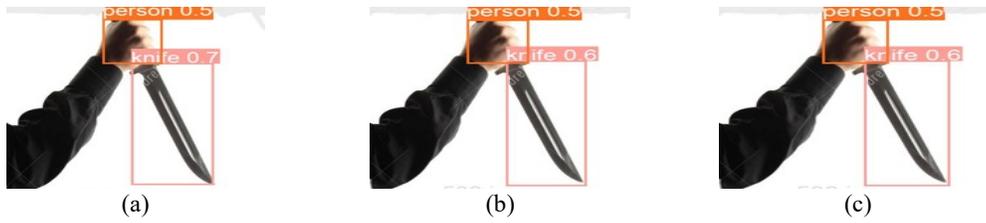

*Figure 5: YOLOv5 Ensemble Model detections*
$Y_{so}$ & $Y_{mo}$ (b) $Y_{mo}$ & $Y_{lo}$ (c) $Y_{lo}$ & $Y_{so}$

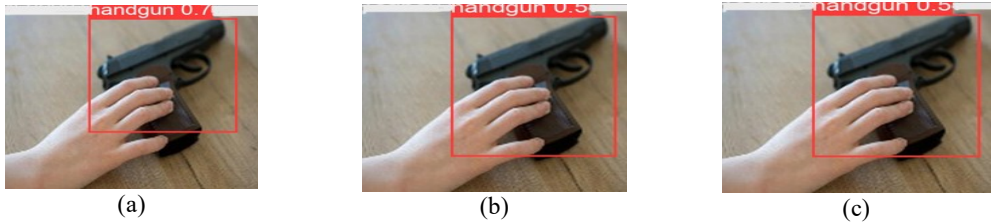

*Figure 6: Sample YOLOv5 Pruning detections.*
*(a) $Y_{so}$ pruning (b) $Y_{mo}$ pruning (c) $Y_{lo}$ pruning*

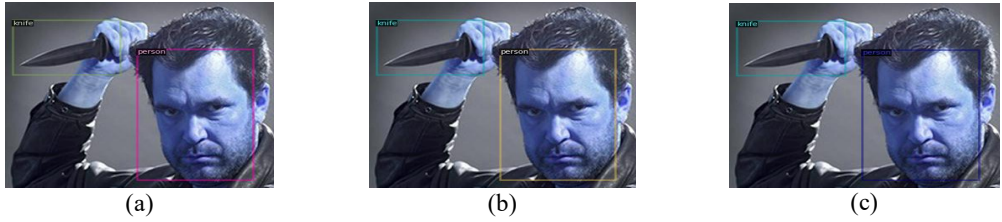

*Figure 7: Faster RCNN Resnet 50 Detections*
*(a) FPN (b) $C_4$ (c) $DC_5$*

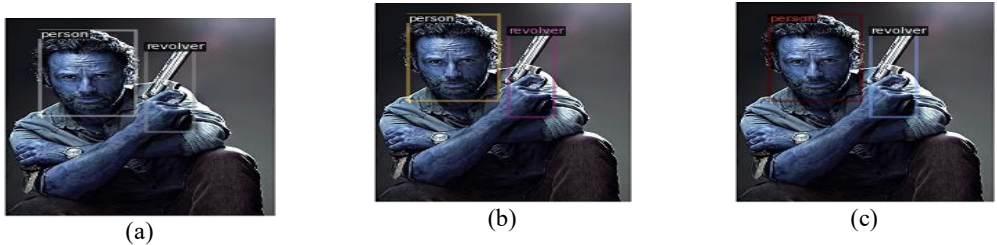

*Figure 8: Faster RCNN Resnet 101 Detections*
*(a) FPN (b) $C_4$ (c) $DC_5$*

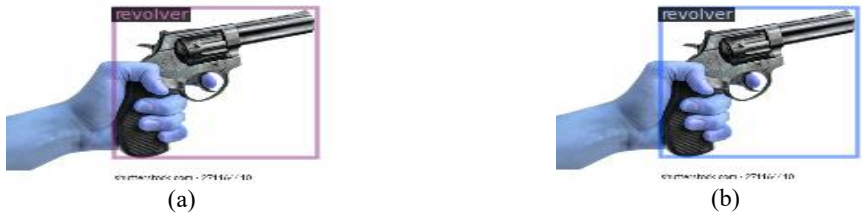

*Figure 9: Faster RCNN Retinanet Detections*
*(a) R-50- FPN (b) R-101-FPN*

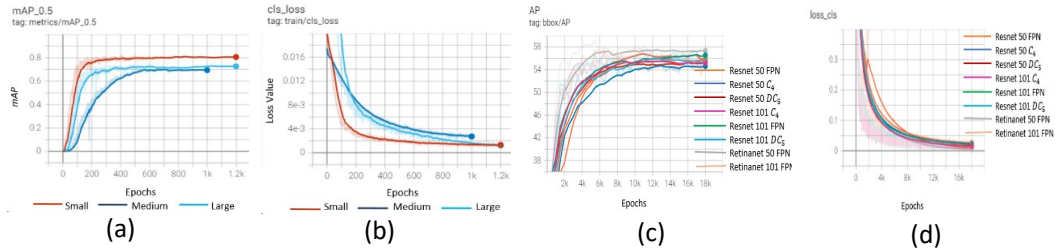

*Figure 10: Metric Graphs: YOLO results (a) Mean Average Precision (mAP) of $Y_{so}, Y_{mo}, Y_{lo}$ models. (b) YOLO training classification Loss (c) Faster RCNN Average Precision (AP) of Resnet (50,101) and Retinanet (50,101) models (d) Faster RCNN training classification loss of Resnet (50,101) and Retinanet (50,101) models.*

# 5. CONCLUSION & AMELIORATION

## 5.1. Threats to validity & Ablation Study

Neural Network Architecture in $YOLOV_5$ family is ablated and customized by mathematical procedure. CSP dark net is replaced by $C_3$ neural layers where initial model training results are not up to the mark. The model was under-fitting because of high bias and low variance. Upon understanding the technicality of each neural layer, there is necessary to change the activation layer from ReLU to Sigmoid Linear Activation, which enhances the model's performance. The main categories in the theory of validity are external and internal validity. This model achieves external validity as it detects not only weapons but also human detection in any public area. The model achieves internal validation by dataset design and model training. Data augmentation techniques, including random images as noise to model that which improves models' robustness. Some of the threats to validity are timing and selection bias. The resultant model can be introduced as a pre-trained model for future model training. The dataset is the homogeneous type which avoids selection bias that serves as a threat to internal validity.

## 5.2. Amelioration work

The experimental results show that *Resnet-50-FPN* achieved the highest average precision in detecting weapons, while $Y_{so}$ showed the highest precision in classifying weapon classes. However, YOLOv5 ensemble model ($Y_{so}, Y_{mo}, Y_{lo}$) showed the highest recall and mean average precision in detection and classification. The developed algorithm succeeded in detecting and classifying different classes. Resnet 50 *R-50-FPN* and $Y_{so}$ could be considered as supporting models to detect weapons in real-time scenarios since these models achieve the trade-off between accuracy and real-time performance. Further extension of the research can include increasing the dataset that comprises different weapon classes and thermal images of weapons for supporting night vision detection. The current supervised model involves a lot of manual annotation which can be improved by implementing *self-supervised algorithms* that reduce human labelling.

**Authors**

Akhila Kambhatla

is currently pursuing the Ph.D. degree in Southern Illinois University, Carbondale, USA. Her research interest includes machine learning, deep learning, swarm intelligence, object detection and computer vision. Her interests not limited to federated Learning, Adversarial Neural Networks and Block Chain Technology.

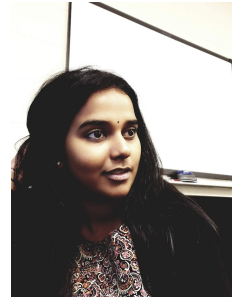

Khaled R Ahmed

is currently an assistant professor at Southern Illinois University, Carbondale. His research interests include machine learning, deep learning, computer vision, high-performance computing, distributed and parallel computing, and big data. He published over 63 articles in journals and proceedings and was involved as PI and Co-I in about 8 funded research projects. He edited four books in Peer-to-peer, Wireless sensor networks, advances in Bigdata, blockchain and deep learning. He was an associate professor of computer science at King Faisal University till 2018.

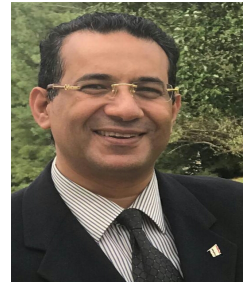